\documentclass{bmvc2k}

\usepackage{ifpdf}
\usepackage[normalem]{ulem}
\useunder{\uline}{\ul}{}
\usepackage{amsmath}

\usepackage{algorithmic}

\usepackage{array}

\usepackage{dblfloatfix}

\usepackage{url}
\usepackage{floatrow}
\usepackage{graphicx}
\usepackage{algorithm2e}
\usepackage{multirow}

\newcommand{\headcell}[2][c]{%
  \begin{tabular}[#1]{@{}c@{}}#2\end{tabular}}

\usepackage{sidecap}

\title{Improved Visual Relocalization by Discovering Anchor Points}

\addauthor{Soham Saha$^{\text{1,~}}$}{soham.saha@research.iiit.ac.in}{2}
\addauthor{Girish Varma}{girish.varma@iiit.ac.in}{1}
\addauthor{C V Jawahar}{jawahar@iiit.ac.in}{1}

\addinstitution{
Kohli Center for Intelligent Systems, Center for Visual Information Technology,
IIIT Hyderabad, India.}
\addinstitution{
Flipkart Internet Private Limited,\\ Bangalore, India.
}

\runninghead{Saha, Varma and Jawahar}{Visual Relocalization using Anchor Points}

\begin{document}

\maketitle

\begin{abstract}
We address the visual relocalization problem of predicting the location and camera orientation or pose (6DOF) of the given input scene. We propose a method based on how humans determine their location using the visible landmarks. We define anchor points uniformly across the route map and propose a deep learning architecture which predicts the most relevant anchor point present in the scene as well as the relative offsets with respect to it. The relevant anchor point need not be the nearest anchor point to the ground truth location, as it might not be visible due to the pose. Hence we propose a multi task loss function, which discovers the relevant anchor point, without needing the ground truth for it. We validate the effectiveness of our approach by experimenting on Cambridge Landmarks (large scale outdoor scenes) as well as 7 Scenes (indoor scenes) using various CNN feature extractors. Our method improves the median error in indoor as well as outdoor localization datasets compared to the previous best deep learning model known as PoseNet (with geometric re-projection loss)  using the same feature extractor.  We improve the median error in localization in the specific case of Street scene, by over 8m.

\end{abstract}

\section{Introduction}
\label{sec:intro}

The visual relocalization problem is an essential component of many practical systems like autonomous  navigation, augmented reality, drone navigation etc. Although GPS sensors could be used for localization in these applications, it is often noisy and will not work in indoor environments.  Hence an independent source of location information is essential for safety as well as applicability in a wide variety of environments. Though location information could be accurately estimated using 3D point cloud data, gathering and processing such data can be expensive and slow. The visual relocalization problem is defined as estimating the location as well as camera pose, given just the observed camera frame, without using any other sensor data.

Early solutions modeled this problem as an image retrieval problem. However, these solutions required features for a large collection of images to be stored. Also, a nearest neighbour search was needed at test time. Thus the memory and runtime increased proportionately with the dataset size. The most accurate approaches require 3D point cloud data of the region \cite{active_search}, which is expensive to gather and process. PoseNet proposed to model this problem directly as a regression problem and used a deep neural network as the image feature extractor \cite{kendall2015posenet}. Extensions to this work aim at modeling the uncertainty in estimating the location and pose by using a Bayesian neural network \cite{kendall2017geometric}. Furthermore, a spatial LSTM based approach to regress the pose was found to improve performance \cite{walch2016image}. A more recent approach has been to improve points through a geometric loss function, which tries to minimize the reprojection error \cite{kendall2017geometric}.


\begin{SCfigure}
  \centering
  
  \includegraphics[width=0.48\textwidth]%
    {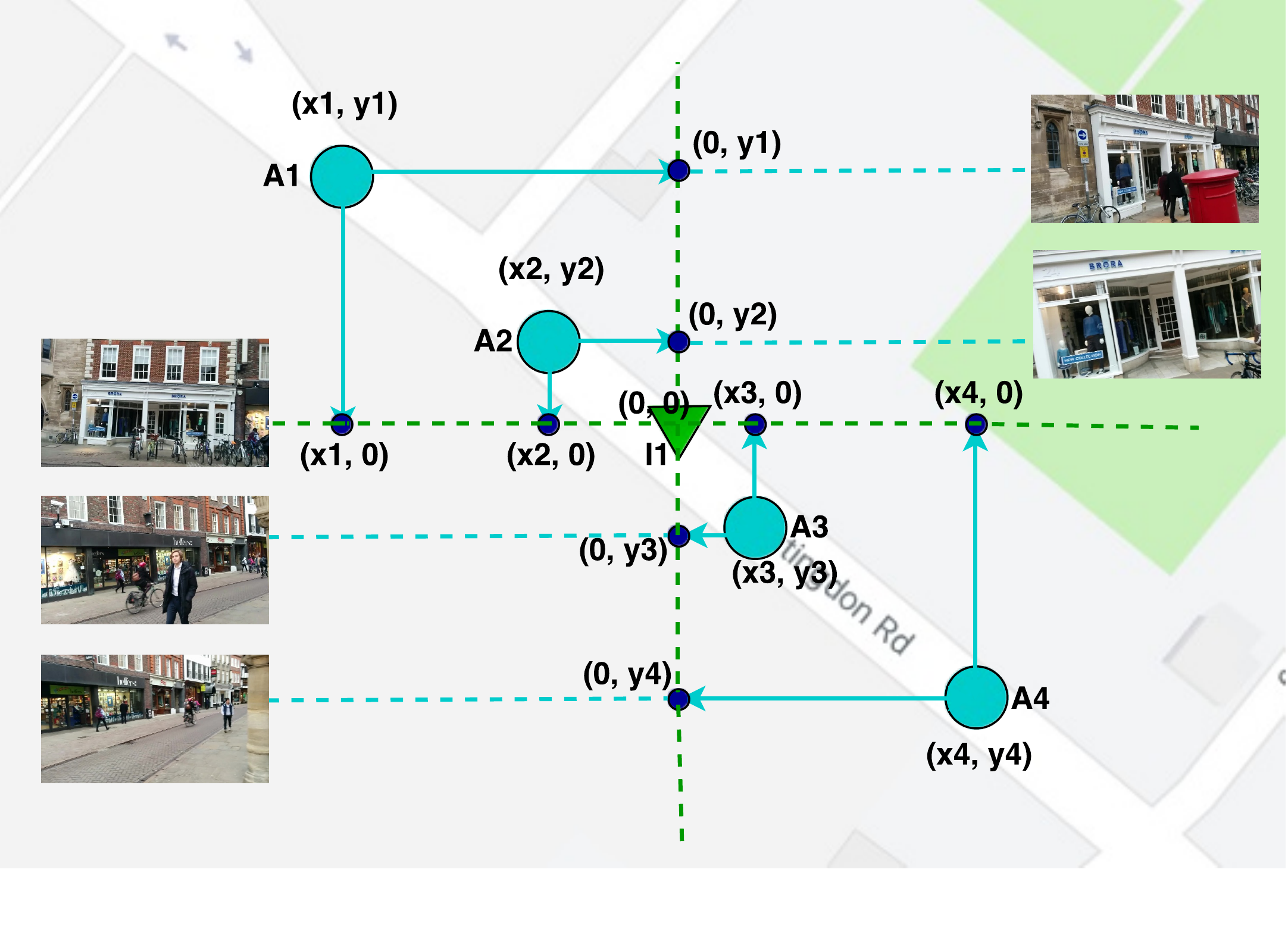}
    \vspace{-2em}
      \caption{
      \small An example of anchor point allocation for a sample road. The blue points (A1-A4) denote the anchor points which are predefined uniformly. The green triangle(I1) is the input frame which denotes the current location. The coordinates of the anchor points are shown relative to the coordinate of the current location (denoted as (0,0)). The nearest anchor point A3 or A2, might not be visible properly from the current location. Our system can find the most relevant anchor point and the relative offsets from it, giving a more accurate prediction.\vspace{-5em}}
      \vspace*{-3em}
\end{SCfigure}

We propose a visual relocalization system inspired by how humans determine location. We typically identify some landmarks which we deem as important. The coordinates of these landmarks are already known. Then, we try to estimate our position by determining the offset to our position relative to the landmark. The landmark we choose need not be the nearest one, as it might not be visible, due to the viewing angles. We refer to these landmarks as anchor points. Inspired by this idea, we propose a system which assigns anchor points relative to which the 6DOF (six degrees of freedom of the 3 spatial and 3 pose coordinates) can be predicted accurately. We do this by modeling the problem as a multi task problem, of classifying the input scene to an anchor point and then finding the offsets relative to that anchor point. However, a direct approach for training such a network requires ground truth of the anchor points for each image. The datasets typically provide only the ground truths of the 6-DOF and not for relative offsets from the anchor points. The anchor point visible in the image need not be the nearest one, because of the pose. Hence we propose a new loss function, which during training, automatically finds the appropriate anchor point relative to which the offsets needs to be regressed in an end to end fashion.

We benchmark our model on an outdoor dataset covering a large area (Cambridge landmarks \cite{kendall2015posenet} covering few 1000 m$^2$) as well as an indoor dataset (7 Scenes \cite{shotton7scenes} covering few m$^2$) suited for small robot navigation. We improve the median error in all the scenes of both the datasets, from the previous best model, PoseNet (with geometric reprojection loss~\cite{kendall2017geometric}), when using the same GoogleNet \cite{googlenet} feature extractor. We achieve \textless1.5m and \textless4$^\circ$ in localization performance in 4 out of the 6 outdoor scenes in Cambridge Landmarks. In the specific case of Street scene, our method improves the median error by over 8m. Furthermore, we localize to within \textless 0.2m for all the indoor scenes of the 7 Scenes dataset which is a significant improvement over the previous deep learning based approaches. We also experiment with various feature extractors like DenseNet \cite{densenet2016} (giving better accuracy) and MobileNet \cite{MobileNet} (giving better runtime performance, while maintaining accuracies). 

\pagebreak
\section{Related Works}
In this section, we briefly survey the literature. For a detailed survey see \cite{visloc_survey}.
\vspace{-1em}
\paragraph{Visual Place Recognition.} 


The visual place recognition task has been traditionally solved by modeling it as an image retrieval problem \cite{dislocation_arandjelovic2014}, \cite{prob_localization_cummins2008}, \cite{visual_repitative_structures_torii2013},
\cite{scene_coordinate_regression2013}. This enables Bag of Words (BOW) and VLAD \cite{VLAD} representations to be used in scalable retrieval approaches. More recently, deep learning models have also been used successfully for creating efficient image representations, which have been combined with retrieval methods \cite{visual_retrieval_dcn}, \cite{multi_scale_dcn2014}, \cite{aggregating_retireval2015}, \cite{particular_tolias2015} \cite{neural_code_babenko2014}. However, retrieval based solutions require  image features for the entire dataset to be stored. Also, a nearest neighbour search needs to be triggered at test time. Thus, the memory and runtime increased proportionately with the data set size. PlaNet \cite{planet2016} modeled the problem as a classification problem for localization at a world scale. However, estimation of a fine-grained 6-DOF localization requires continuous output and discrete methods have not worked so far for this task.
 
In contrast, metric localization (visual localization) techniques approach this as a 2D-3D mapping between the 2D coordinate system of the image space to the 3D coordinate system of the world space. This is typically done by matching the representations of the learnt image \cite{li2010location,choudhary2012visibility,sattler2012improving,svarm2014accurate,li2016worldwide,active_search} and adapting a nearest neighbour approach. The full 6-DOF pose of the camera can be estimated quite precisely. These methods however require a large database of features and efficient retrieval methods, which makes them suffer at test time since most retrieval methods have an additional feature matching latency.

\vspace{-1em}
\paragraph{PoseNet.} PoseNet \cite{kendall2015posenet} addressed this idea by introducing a regression based deep neural network technique to estimate the metric localization parameters. It achieves a combination of strengths of place recognition and localization approaches and can localize without a prior estimate on the pose. This solves the disadvantage of storing the features in memory since they are learnt during training and estimated during testing.

Extensions to this work have been on RGB-D input images for localization \cite{li2017indoor}, learning relative ego-motion of the camera \cite{melekhov2017relative}, improving the context of features used for localization \cite{szegedy2015going}, using video sequence information for localization \cite{kendall2016modelling}, modeling the uncertainly in localization using Bayesian Neural Networks \cite{kendall2016modelling}, exploring a number of loss functions for learning camera pose which are based on geometry and scene re-projection error \cite{kendall2017geometric}.

Although PoseNet and its extensions are robust and scalable, they lag behind some of the traditional methods \cite{active_search} in estimating the pose . In this work, we propose a novel discrete anchor point assignment based regression technique which provides significant improvement over these methods while maintaining the scalability and robustness. We keep intact the advantages of PoseNet and its variants as well as improve over the nearest neighbour based methods, where feature vectors and required to be stored in memory entirely.

\section{Approach}
\label{approach}
\begin{figure}[!tbh]\label{fig:arch}
\centering
\includegraphics[scale=0.18]{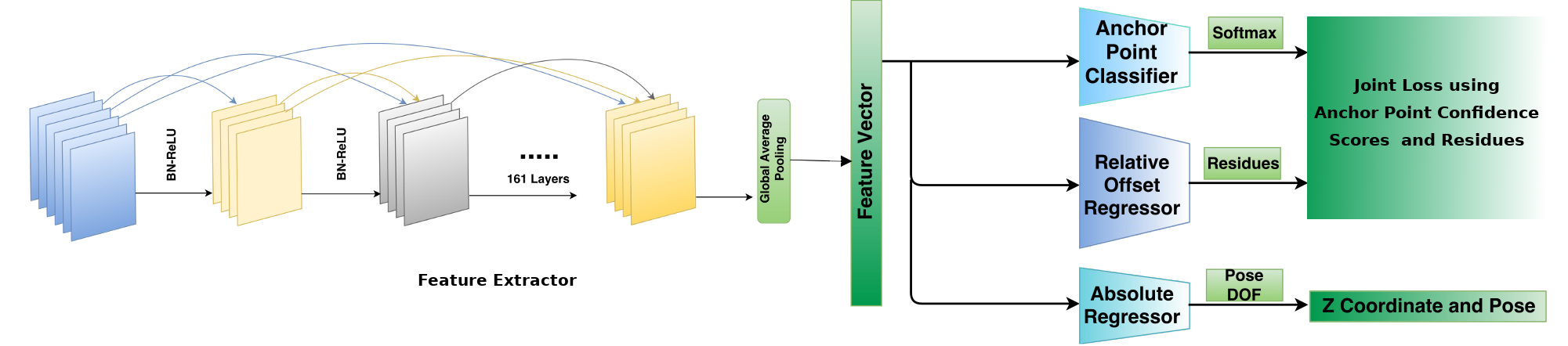}
\caption{\small Block diagram of the proposed architecture. We experiment with different CNN feature extractors including GoogleNet, DenseNet and MobileNet. The CNN feature extractor is followed by a global average pooling layer to get the final feature vector. The feature vector is fed to an anchor point classifier, relative offset regressor and an absolute regressor for pose. \vspace{-1em}}
\vspace{-1em}
\end{figure}

The visual relocalization problem involves inferring the 3D location coordinates and the camera pose (given by 3 angles) from just the image. In the previous works \cite{kendall2016modelling,kendall2017geometric,kendall2015posenet}, a deep neural network based approach was proposed which directly  predicts the 6-DOF by regressing against the ground truth, relative to a global coordinate system. Here, we propose a system which assigns anchor points relative to which the 6-DOF can be predicted accurately. We do this by modeling the problem as a multi-task problem. The first task is of classifying the input camera image to one of the anchor points and the second task is of finding the offsets relative to that anchor point.  We use standard CNN feature extractors for image feature extraction. The feature extractors are typically followed by a Global Average Pooling (GAP) layer in order to get a single vector instead of a feature map. This is illustrated in Figure \ref{fig:arch}. Our proposed method then branches this output into 3 heads.

\vspace{-1em}
\paragraph{Anchor Point Classification.}
Given a route map where visual relocalization is desired, we subdivide the route  into equally spaced intervals. The end points of the intervals are designated as anchor points for the 2D spatial coordinates. Note that the global coordinates of these anchor points are known, since they are fixed a-priori. We then model the problem of finding the most relevant anchor point as an image classification problem. Given an image, the predicted class should be the most prominent anchor point in the image. Note that the most prominent anchor point need not always be the one nearest to the location of the image. Hence, we frame our problem in such a way that we need not have the ground truth information about the prominent anchor point directly as part of the training data set. We  treat the predicted probabilities as a confidence score for each of the anchor points. We combine this information in our loss with the relative offsets. Furthermore, we experiment by considering the nearest anchor point to the image, as the ground truth, and incorporate the corresponding cross entropy loss in our overall loss function. We report results for both cases.

The Anchor Point Classification head is obtained by mapping the global average pooling output of the feature extractor to the number of predefined classes using a Fully Connected (FC) layer. In our case, the number of classes is equal to the number of anchor points defined during preprocessing. The output of the FC layer is then subjected to a softmax classifier to get a probability mapping. However, instead of using these values as probability predictions, we use them as confidence scores in our experiments as will be explained in more detail in Section \ref{loss_func}. 

\vspace{-1em}
\paragraph{Relative Offset Regression.}
Along with anchor point classification, our system also produces the relative offsets from the anchor point. We train these offsets through the regression loss. However, we face the same problem of not having the ground truth for the most relevant anchor point. Hence our system produces the relative offsets with respect to all the anchor points and our model tries to regress these relative offsets. Note that since the global coordinates of all the anchor points are known, the relative offset ground truth information for all the images can also be calculated from the global ground truth information. The output of the GAP layer of the feature extractor is passed through a fully connected neural network layer to obtain the regressing head whose task is to predict the X,Y Coordinate offset values relative to each anchor point defined previously. Thus if there were $N$ anchor points, say, the output of this head would be a $2N$ dimensional vector.
\vspace{-1em}
\paragraph{Absolute Offset Regression for Z and Pose.}
Our approach is motivated from human scene recognition through relative identification. Hence, having the anchor points can be efficient only for the X and Y coordinates and not for the Z and camera angle coordinates. Predicting relative offsets for Z coordinate and pose is not carried out due to the following reasons:
\begin{enumerate}
\item The Z coordinate and the camera pose of a scene is independent of the previous scene. It is thus counter intuitive and does not capture the continuity of the scene like the X,Y coordinates do.
\item Regressing relative offset for the Z coordinate as well as the pose complicates the regression task further thereby leading to inferior results as we found out in our experiments.
\end{enumerate} 
Hence, our model separately predicts the the remaining 4-DOF in accordance with the global coordinate system. The third and final head is assigned the task of predicting the absolute values of the Z coordinate and the remaining DOF values for determining the camera pose. Therefore, the GAP layer of the feature extractor is mapped to a 5 dimensional vector in this case. This head also consists of a fully connected layer for regressing the absolute values of the remaining DOF (i.e Z coordinate and pose).

\subsection{Loss Function}
\label{loss_func}

We train our model with a loss function which is the sum of 3 components. Let us denote the output of the anchor point classification head to be $\hat C$, which gives confidence scores for each of the $N$ anchor points, for the current input image. $X,Y$ denotes the vectors of dimension $N$ of ground truth offsets of the frame in the horizontal plane, with respect to each of the $N$ anchor points and $Z$, the vertical distance. The ground truths $X,Y$, though not provided with the dataset, can easily be computed as a preprocessing step, by changing the origin of the coordinate system to the respective anchor points. $P$ denote the three angles of the camera. Let $\hat X, \hat Y, \hat Z, \hat P$, be the predictions of the model.

Since we do not have ground truths for the most relevant anchor point in the image, we cannot train the anchor point classification head with a cross entropy loss. However, we formulate a joint loss function that uses $\hat C$ and the offsets $\hat X, \hat Y$ from each of anchor points. 
We weight the squared loss of offsets  with confidence scores of the respective anchor points. The resultant loss function therefore looks like the following:
\begin{equation}
\label{loss1}
\sum_i\left[\left(X_i - \hat{X}_i\right)^2 + \left(Y_i - \hat{Y}_i\right)^2\right]\hat C_i
\end{equation}
This is motivated by the fact that the accuracy is determined by the offset of the most relevant anchor point. If an anchor point is completely irrelevant ($\hat C_i = 0$), then we do not need the corresponding offset predictions.

Now, we include the predictions of the absolute regressor which is responsible for predicting the Z coordinate values and the remaining angular DOF for determining the pose. In order to do this, add the following component to the loss function:
\begin{equation}
\label{loss2}
\sum_i\left[\left(Z_i - \hat{Z}_i\right)\right]^2 + \left\| P_i - \frac{\hat{P_i}}{\left\|\hat{P_i}\right\|}\right\|^2
\end{equation}

Additionally, we also experiment with adding a cross entropy loss with the ground truth being nearest anchor point.
\begin{equation}
\label{loss3}
\text{Confidence loss} = H\left(C_i,\hat{C}_i\right)
\end{equation}

All the 3 components of our loss function mentioned in equations \ref{loss1}, \ref{loss2}, \ref{loss3} are then assigned weights (hyperparameter) and the summed up to get the overall loss for our task. Therefore, we now have the following resultant loss function:
\begin{multline}
\label{total_loss}
\alpha_1 H(C_i,\hat{C}_i) + \alpha_2\sum_i\left[\left(X_i - \hat{X}_i\right)^2 + \left(Y_i - \hat{Y}_i\right)^2\right]C_i  + \alpha_3\sum_i\left[\left(Z_i - \hat{Z}_i\right)\right]^2 + \left\|P_i - \frac{\hat{P_i}}{\left\|\hat{P_i}\right\|}\right\|^2
\end{multline}
where $\alpha_1, \alpha_2, \alpha_3$ are weights assigned to the separate components of the loss function.

%
%

\vspace{-1em}
\section{Dataset and Experiments}

We benchmark our method on an outdoor and as well as indoor localization dataset. A summary of the datasets can be found in Table \ref{tab:datasets}.  We rescale all the images to $224 \times 224$  since our convolutional feature extractors are trained on those dimensions.

As described in Section \ref{approach}, we initialize our feature extractor (DenseNet architecture) pretrained on the Imagenet 1K dataset trained on the image classification task. This gives better performance for outdoor scenes than using a network pre-trained on the Places dataset as shown in earlier work by Kendall et.al \cite{kendall2015posenet}. 

\vspace{-0.5em}
\begin{table*}[!tbh]
\centering
\caption{\small Summary of localization datasets used for benchmarking.} 
\scalebox{0.65}{
\begin{tabular}{|l|r|c|c|c|c|c|c|c|}
\hline
{Dataset}       & \begin{tabular}[c]{@{}c@{}}Type \end{tabular} & \begin{tabular}[c]{@{}c@{}}Scale\end{tabular} & \begin{tabular}[c]{@{}c@{}}Imagery\end{tabular} & \begin{tabular}[c]{@{}c@{}}Scenes\end{tabular} & \begin{tabular}[c]{@{}c@{}}Train Images\end{tabular} & \begin{tabular}[c]{@{}c@{}}Test Images\end{tabular} & \begin{tabular}[c]{@{}c@{}}Spatial Area\end{tabular} \\ \hline 
Cambridge Landmarks \cite{kendall2015posenet}  & Outdoor & Street & Mobile phone camera  & 6 & 8,380 & 4,841 & 100 $\times$ 500 m  \\ \hline
7 Scenes \cite{shotton7scenes}   & Indoor & Room & RGB-D sensor (Kinect)  & 7  & 26,000 & 17,000 & 4 $\times$ 3 m   \\ \hline

\end{tabular}}
\vspace{-1em}
\label{tab:datasets}
\end{table*} 
\vspace{-0.5em}

\begin{figure*}[!tbh]
\vspace{-1em}
\centering
\includegraphics[scale=0.24]{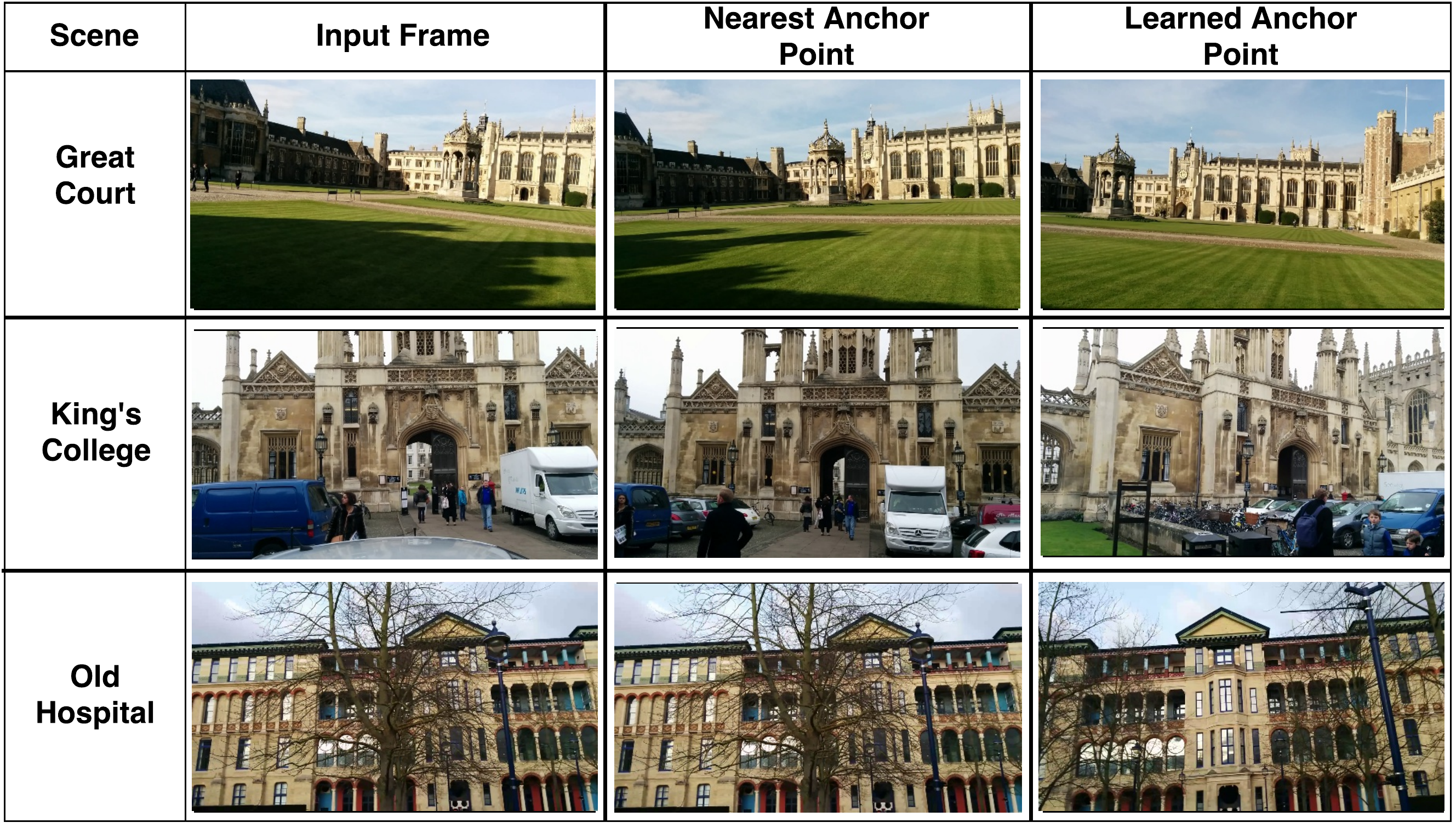}
\hspace{0.2cm}
\includegraphics[scale=0.24]{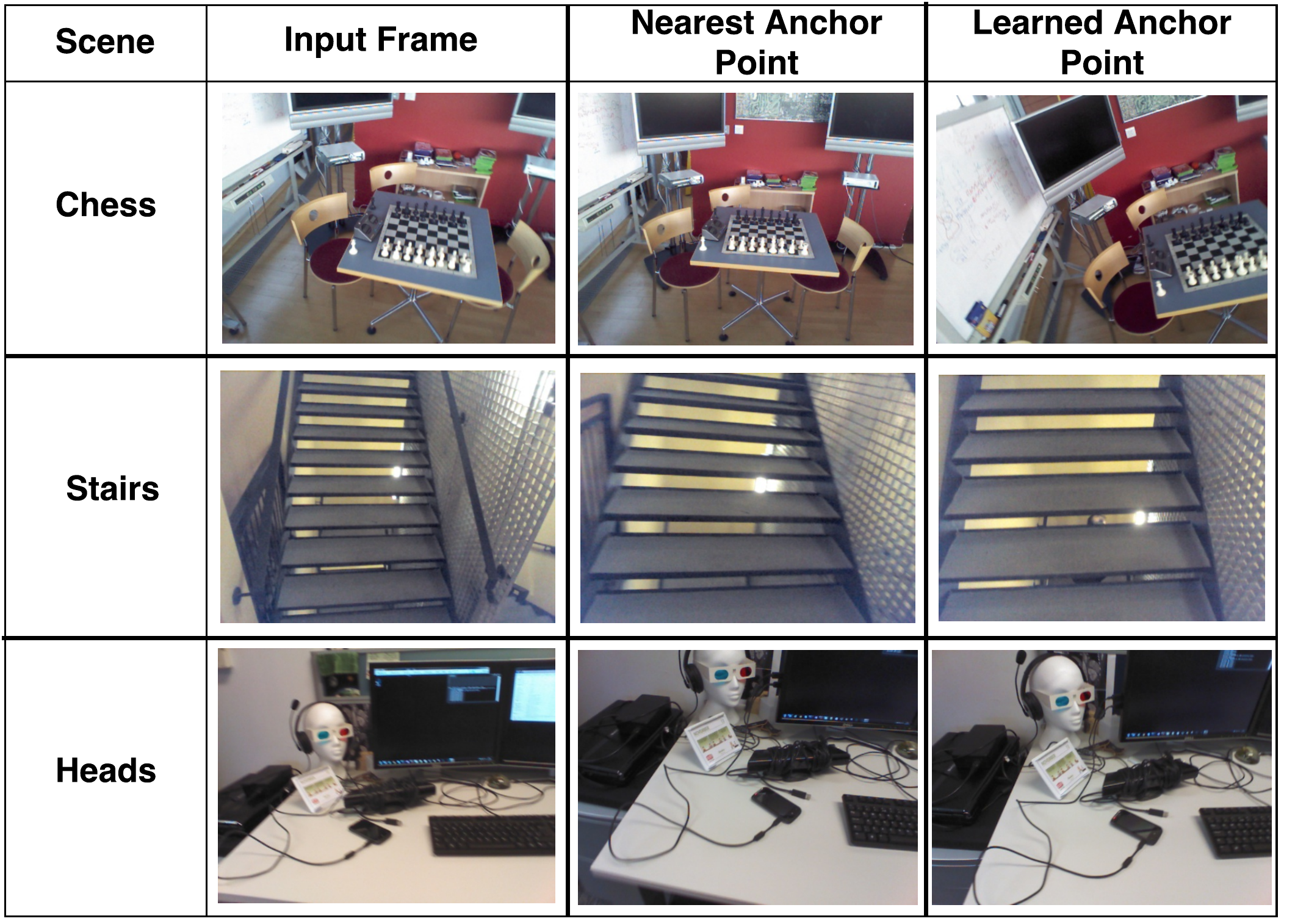}
\vspace{-1em}
\caption{ \small (Left) We contrast the nearest anchor point and the learned anchor point for an input query for the Cambridge dataset. Note that for the Old Hospital scene, the relevant anchor point learned is not blocked by a tree while the nearest anchor point is. Generalizing, learned anchor points gives a clear view of a landmark as compared to the nearest anchor point, validating our approach of discovering the relevant anchor point. (Right) Learned anchor points from the 7 Scenes dataset. In this case we observe a more zoomed in version of the input image is learned as the reference anchor point.}

 \label{fig:anchor_points}

\end{figure*}

As a pre-processing step, we first divide the scene space into several anchor points by uniformly selecting every n-th frame from among all the frames in a scene video. The distance of x,y coordinates for each image is then calculated relatively from each of these anchor points. This is the ground truth information for the Relative Offset Regressor. We illustrate some of the selected anchor points for 3 scenes, King's College, Shop Facade and Street in Figure \ref{fig:anchor_points}.

Our proposed loss function has 3 separate components. We assign separate weights to each of these components, all of which are hyperparameters, and are subject to be optimized through grid-search or any other hyperparameter optimization technique. We found specific ranges for these weights for outdoor scene localization task. The first part of the loss function is the cross-entropy calculated between the classifier output and the nearest anchor point to an input scene, which acts as its label. For this classification loss, we use a weight value ranging between 1-3 across all the scenes. The second component of the loss function is the Relative Offset (Translation) Regressor which predicts the relative distance of the x,y coordinates of the input scene from each of the anchor points. As a weight for this component, we use a value ranging between 4-30. Finally, the pose regressor predicts the remaining 5-DOF and a weight of 0.1-2 is assigned to it. 

We use a learning rate varying from 0.00005 to 0.0005 for the scenes in the dataset. The Adam optimizer is used for optimization with a scheduler to decay the learning rate by half after every 30th epoch. The evaluation of accuracy is a subjective procedure. We consider a prediction for the scene to be accurate if the prediction for position is \textless 2m and the corresponding pose is \textless 5$^\circ$. Other thresholds for accuracy in evaluation will naturally produce variations in performance calculation.


\section{Results and Discussions}
For validating the effectiveness of our approach, we first compare the median errors for 6-DOF pose in the spatial and angular dimensions using our approach, as well as the previous best PoseNet model \cite{kendall2015posenet} . For fairness of comparison, we experiment with using the same GoogleNet \cite{googlenet} feature extractor as the PoseNet result. Secondly, we also experiment with newer feature extractors like DenseNet \cite{densenet2016} and MobileNet \cite{MobileNet} to drastically improve the accuracy and runtime speed. This also demonstrates that our approach generalizes to a variety of feature extractors. Furthermore, we provide a more detailed analysis of anchor points as well as qualitative results.
\vspace{-1em}
\paragraph{Comparison with PoseNet on GoogleNet.}
We compare our method with PoseNet. We use the same GoogleNet feature extractor used in the PoseNet results, so that we can observe the improvement due to our approach rather than using a improved feature extractor. We report results without the cross entropy loss applied to the classification head since letting the network discover relevant the anchor points gave better results. We report the median error for all the scenes in the Cambridge Landmarks and the 7 Scenes  datasets  in Table \ref{table2}.  Also, it can be seen from Table \ref{table2},  that we are doing better than the previous best method of PoseNet (with geometric reprojection loss) \cite{kendall2017geometric} in all the scenes, and improve the localization median error by around 8m in the Street scene. We also improve upon the previous best performance for each of the indoor scenes in the 7 Scenes dataset.

\begin{table*}[!tbh]
\vspace{-0.5em}
\caption{\small Comparison of median error for 6-DOF pose. As reported in the last  column, our model performs better than the best deep learning model PoseNet \cite{kendall2017geometric}, making the gap with active search methods \cite{active_search} (which uses 3D point cloud data unlike us) lesser. Use of improved feature extractors like DenseNet reduces the errors further.}

\centering

\scalebox{0.5}{
\begin{tabular}{|l|r|c|c|c|c|c|c|c|}
\hline
{\ul Scene}  & \headcell{Area or \\ Volume} & \headcell{Active Search \\ (SIFT) \cite{active_search}}  & \headcell{Posenet\\ Spatial LSTM\cite{walch2016image}} & \headcell{Posenet \\ sigma$^2$ weight \cite{kendall2017geometric}} & \headcell{Posenet\\ Geom. Rep. \cite{kendall2017geometric}} & \headcell{Ours (DenseNet) \\ (cross entropy)} & \headcell{Ours (DenseNet)\\ (w/o cross entropy)} & \headcell{\bf Ours (GoogleNet)\\ \bf (w/o cross entropy)}  \\ 
\hline
Great Court & 8000$m^2$  & -                                                        & -                                                              & 7.00$m$, 3.65$^\circ$                                                            & 6.83$m$, 3.47$^\circ$                                                              & 5.85$m$, 3.61$^\circ$                                                           & 4.64$m$, 3.42$^\circ$ & \textbf{5.89$m$, 3.53$^\circ$} \\ 
\hline
King's College    & 5600$m^2$                                                    & 0.42$m$, 0.55$^\circ$                                                                                                     & 0.99$m$, 3.65$^\circ$                                                    & 0.99$m$, 1.06$^\circ$                                                            & 0.88$m$, 1.04$^\circ$                                                              & 0.55$m$, 0.97$^\circ$                                                  & 0.57$m$, 0.88$^\circ$
& \textbf{0.79$m$, 0.95$^\circ$}\\ 
\hline
Old Hospital      & 2000$m^2$                                                    & 0.44$m$, 1.01$^\circ$                                                                                                     & 1.51$m$, 4.29$^\circ$                                                    & 2.17$m$, 2.94$^\circ$                                                            & 3.20$m$, 3.29$^\circ$                                                              & 1.45$m$, 3.16$^\circ$                                                           & 1.21$m$, 2.55$^\circ$
& \textbf{2.11$m$, 3.05$^\circ$}\\ 
\hline
Shop Facade       & 875$m^2$                                                     & 0.12$m$, 0.40$^\circ$                                                                                                     & 1.18$m$, 7.44$^\circ$                                                    & 1.05$m$ 3.97$^\circ$                                                             & 0.88$m$, 3.78$^\circ$                                                              & 0.49$m$, 2.42$^\circ$                                                  & 0.52$m$, 2.27$^\circ$
& \textbf{0.77$m$, 3.25$^\circ$}\\ 
\hline
St. Mary's Church & 4800$m^2$                                                    & 0.19$m$, 0.54$^\circ$                                                                                                     & 1.52$m$, 6.68$^\circ$                                                    & 1.49$m$, 3.43$^\circ$                                                            & 1.57$m$, 3.32$^\circ$                                                              & 1.12$m$, 2.84$^\circ$                                                           & 1.04$m$, 2.69$^\circ$
& \textbf{1.22$m$, 3.02$^\circ$}\\ 
\hline
Street            & 50000$m^2$                                                   & 0.85$m$, 0.83$^\circ$                                                                                                                       & -                                                              & 20.7$m$, 25.7$^\circ$                                                            & 20.3$m$, 25.5$^\circ$                                                              & 8.19$m$, 25.5$^\circ$                                                           & 7.86$m$, 24.2$^\circ$
& \textbf{11.8$m$, 24.3$^\circ$ }  \\ 
\hline
\multicolumn{8}{|l|}{}    
\\ \hline
Chess             & 6$m^2$                                                      & 0.04$m$, 1.96$^\circ$                                                                                                     & 0.24$m$, 5.77$^\circ$                                                    & 0.14$m$, 4.50$^\circ$                                                            & 0.13$m$, 4.48$^\circ$                                                              & 0.06$m$, 3.95$^\circ$                                                           & 0.06$m$, 3.89$^\circ$
& \textbf{0.08$m$, 4.12$^\circ$}\\ 
\hline
Fire              & 2.5$m^2$                                                    & 0.03$m$, 1.53$^\circ$                                                                                                      & 0.34$m$, 11.9$^\circ$                                                    & 0.27$m$, 11.8$^\circ$                                                            & 0.27$m$, 11.3$^\circ$                                                              & 0.16$m$, 10.4$^\circ$                                                                     & 0.15$m$, 10.3$^\circ$
& \textbf{0.16$m$, 11.1$^\circ$}\\ 
\hline
Head              & 1$m^2$                                                       & 0.02$m$, 1.45$^\circ$                                                                                                     & 0.21$m$, 13.7$^\circ$                                                    & 0.18$m$, 12.1$^\circ$                                                            & 0.17$m$, 13.0$^\circ$                                                              & 0.08$m$, 10.7$^\circ$                                                                     & 0.08$m$, 10.9$^\circ$
& \textbf{0.09$m$, 11.2$^\circ$}\\ 
\hline
Office            & 7.5$m^2$                                                     & 0.09$m$, 3.61$^\circ$                                                                                                     & 0.30$m$, 8.08$^\circ$                                                    & 0.20$m$, 5.77$^\circ$                                                            & 0.19$m$, 5.55$^\circ$                                                              &  0.11$m$, 5.24$^\circ$                                                                     & 0.09$m$, 5.15$^\circ$
& \textbf{0.11$m$, 5.38$^\circ$}\\ 
\hline
Pumpkin           & 5$m^2$                                                       & 0.08$m$, 3.10$^\circ$                                                                                                     & 0.33$m$, 7.00$^\circ$                                                    & 0.25$m$, 4.82$^\circ$                                                            & 0.26$m$, 4.75$^\circ$                                                              &  0.11$m$, 3.18$^\circ$                                                                    & 0.10$m$, 2.97$^\circ$
& \textbf{0.14$m$, 3.55$^\circ$}\\ 
\hline
Red Kitchen       & 18$m^2$                                                      & 0.07$m$, 3.37$^\circ$                                                                                                     & 0.37$m$, 8.83$^\circ$                                                    & 0.24$m$, 5.52$^\circ$                                                            & 0.23$m$, 5.35$^\circ$                                                              & 0.08$m$, 4.83$^\circ$                                                                     & 0.08$m$, 4.68$^\circ$
& \textbf{0.13$m$, 5.29$^\circ$}\\ 
\hline
Stairs            & 7.5$m^2$                                                     & 0.03$m$, 2.22$^\circ$                                                                                                     & 0.40$m$, 13.7$^\circ$                                                    & 0.37$m$, 10.6$^\circ$                                                            & 0.35$m$, 12.4$^\circ$                                                              & 0.13$m$, 10.1$^\circ$                                                           & 0.10$m$, 9.26$^\circ$
& \textbf{0.21$m$, 11.9$^\circ$}\\ 
\hline
\end{tabular}}

\label{table2}
\vspace{-1em}
\end{table*}
\vspace{-2em}

\paragraph{Improved Accuracy using DenseNet and Abalation study.}

We then experiment with the state of the art CNN feature extractors. DenseNet \cite{densenet2016} is known to give improved accuracies for image classification. In Table \ref{table2}, we report the results of our proposed method with the DenseNet feature extractor and compare with all the previous methods. We report the results with and without the cross entropy term (see Section \ref{loss_func}). For the Cambridge Landmarks,  5 out of the 6 scenes, the model performed better without the cross entropy loss, validating our approach of discovering the appropriate anchor point relative to which offsets needs to be calculated.

\begin{SCtable}\setlength{\belowcaptionskip}{-4em}
\centering
\caption{\small We report the mean and median distances for each scene of the Cambridge Landmarks dataset using the DenseNet feature extractor. The accuracy is calculated by considering the orientation localization within 2m and the pose localization with 5$^\circ$ to be a correct prediction. \vspace{-2em}}
\scalebox{0.6}{
\begin{tabular}{|l|c|c|c|c|}
\hline
{\ul \textbf{Scene}}                                                 & \textbf{\begin{tabular}[c]{@{}c@{}}{\ul Mean} \\ {\ul Distance}\end{tabular}} & \textbf{\begin{tabular}[c]{@{}c@{}}{\ul Median} \\ {\ul Distance}\end{tabular}} & \textbf{\begin{tabular}[c]{@{}c@{}}{\ul Accuracy}\\ {\ul (\textless2m,\textless5$^\circ$)}\end{tabular}} & \textbf{\begin{tabular}[c]{@{}c@{}}{\ul \#Anchor} \\ {\ul Points}\end{tabular}} \\ \hline
\textbf{Great Court}                                                 & 10.48m                                                             & 5.85m                                                               & 69.51 \%                                                                                & 154                                                                 \\ \hline
\textbf{King's College}                                              & 0.76m                                                             & 0.57m                                                               & 93.52 \%                                                                                & 122                                                                 \\ \hline
\textbf{Old Hospital}                                                & 2.93m                                                             & 1.45m                                                               & 85.94 \%                                                                               & 110                                                                 \\ \hline
\textbf{Shop Facade}                                                 & 0.72m                                                             & 0.52m                                                               & 93.76 \%                                                                               & 33                                                                  \\ \hline
\textbf{\begin{tabular}[c]{@{}c@{}}St. Mary's\\ Church\end{tabular}} & 1.62m                                                             & 1.12m                                                               & 88.95 \%                                                                               & 146                                                                 \\ \hline
\textbf{Street}                                                      & 17.45m                                                            & 9.89m                                                               & 11.26 \%                                                                              & 201                                                                 \\ \hline
\end{tabular}}
\vspace{-2em}
\label{table1}

\end{SCtable}

In Table \ref{table1}, we also report the accuracy, the mean and the median distance localized to in orientation, and the median camera angle pose localization for each of the 6 scenes of the Cambridge Landmarks dataset. It can be seen from Table \ref{table1}, that the accuracy for the Street scene is 11.26\%.  The Street scene is the most challenging scene from among the used scenes for the visual localization task since it covers more than $50000m^2$ in area and also contains multiple landmarks. We achieve state-of-the-art performance for translation and pose on the Street scene as well. Our method is able to significantly improve the translation and orientation localization for this particular scene from  20.3m to 7.86m (see Table \ref{table2}) which is a considerable improvement. 

Next, we  perform an experiment with  DenseNet as the feature extractor connected to a 6-DOF regressor directly. This is a sort of control for our proposed method and we do not use an anchor point classifier here. We provide a comparison of our proposed method and directly using the regressor in the Cambridge Landmarks dataset. It is observed that our proposed method is able to perform much better than simply regressing the DOF from the DenseNet feature extractor. The results are summarized in Table \ref{table3}. As can be seen our method gives lesser errors. This verifies that the improvement in median errors is indeed happening due to our approach and not due to the feature extractors only. 

\begin{table}[!tbh]

\centering
\caption{\small Comparison between the DenseNet feature extractor followed by a simple regressor and our proposed method. }
\scalebox{0.75}{
\begin{tabular}{|l|c|c|c|c|}
\hline
\textbf{Scene}          & \textbf{\begin{tabular}[c]{@{}c@{}}Median Dist. \\ (DenseNet + DOF \\ Regressor)\end{tabular}} & \textbf{\begin{tabular}[c]{@{}c@{}}Median Dist. \\ (Our method)\end{tabular}} & \textbf{\begin{tabular}[c]{@{}c@{}}Accuracy\\ (DenseNet + \\ DOF Regressor)\end{tabular}} & \textbf{\begin{tabular}[c]{@{}c@{}}Accuracy\\ (Our Method)\end{tabular}} \\ \hline
\textbf{Shop Facade}    & 1.32m                                                                                          & \textbf{0.52m}                                                                & 82.64\%                                                                                   & \textbf{93.76\%}                                                         \\ \hline
\textbf{King's College} & 1.45m                                                                                          & \textbf{0.57m}                                                                & 81.80\%                                                                                   & \textbf{93.52\%}                                                         \\ \hline
\end{tabular}}
\label{table3}
\vspace*{-1em}
\end{table}

\vspace{-2em}

\paragraph{Runtime Analysis.}

The MobileNet \cite{MobileNet} feature extractor is known to be significantly faster while maintaining accuracies. We illustrate the results when using it in Table \ref{tab:comparison}. For some scenes in Cambridge Landmarks dataset, we empirically compare the performance and efficiency of 3 popular feature extractors, namely GoogleNet, DenseNet and MobileNet, when used with our proposed method. We observe from Table \ref{tab:comparison} that MobileNet provide a considerable better runtime performance, while still giving errors lesser than GoogleNet.
\vspace{-0.5em}

\begin{table*}[!tbh]
\centering
\caption{\small Comparison of different feature extractors on the basis of performance and FLOPs using our proposed anchor point based method for visual relocalization on the Cambridge landmarks dataset.}
\label{tab:comparison}
\scalebox{0.88}{
\begin{tabular}{|l|c|c|c|c|c|c|}
\hline
\textbf{Scene} & \multicolumn{2}{c|}{\textbf{\begin{tabular}[c]{@{}c@{}}DenseNet\\ (Feature Extractor)\end{tabular}}} & \multicolumn{2}{c|}{\textbf{\begin{tabular}[c]{@{}c@{}}GoogleNet\\ (Feature Extractor)\end{tabular}}} & \multicolumn{2}{c|}{\textbf{\begin{tabular}[c]{@{}c@{}}MobileNet\\ (Feature Extractor)\end{tabular}}} \\ \hline
               & \textbf{Performance}                            & \textbf{FLOPs}                                     & \textbf{Performance}                             & \textbf{FLOPs}                                     & \textbf{Performance}                             & \textbf{FLOPs}                                     \\ \hline
Kings College  & 0.57$m$, 0.88$^\circ$                                     & \multirow{2}{*}{5998 M}                            & 0.79$m$, 0.95$^\circ$                                      & \multirow{2}{*}{760 M}                             & 0.67$m$, 0.94$^\circ$                                      & \multirow{2}{*}{569 M}                             \\ \cline{1-2} \cline{4-4} \cline{6-6}
Shop Facade    & 0.52$m$, 2.27$^\circ$                                     &                                                    & 0.77$m$, 3.25$^\circ$                                      &                                                    & 0.60$m$, 2.31$^\circ$                                      &                                                    \\ \hline
\end{tabular}}
\vspace*{-1em}
\end{table*}
\vspace{-2em}


\paragraph{Analysis of Anchor Points and Qualitative Results.}

\begin{SCfigure}\setlength{\belowcaptionskip}{-3em}
\centering
\includegraphics[scale=0.22]{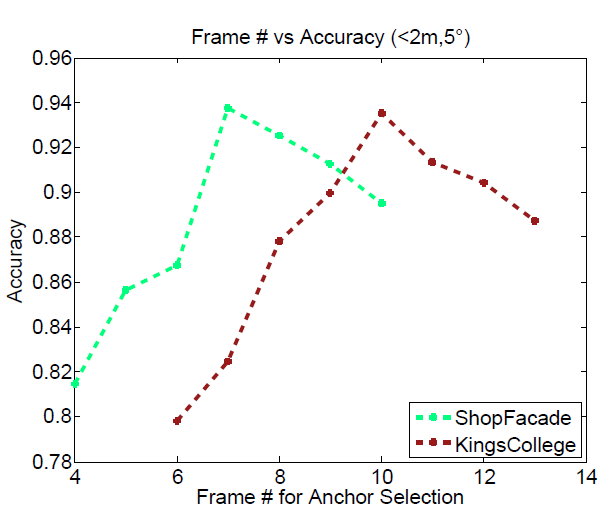}
\includegraphics[scale=0.22]{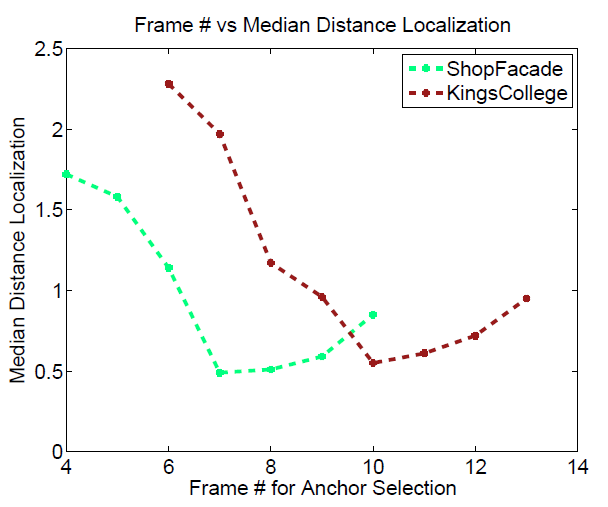}
\caption{\small Plots showing how accuracy and median distance varies with different choices of frames interval between anchor points. We chose the optimum frame number for the dataset for anchor point selection.}
 \label{fig:frame}
 \vspace{-1em}
\end{SCfigure}

An important hyperparameter for our approach is the number of frames between consecutive anchor point which is required for assigning anchor points as a preprocessing step. The outcome for this selection also determines the number of classes the classifier should predict since it is equal to the number of anchor points. We therefore plot the behavior of the median distance localization for translation and the overall accuracy, with the varying frame number selection (say $k$). This means that for a particular scene, we select every $k^{\text{th}}$ frame. We observe in Figure \ref{fig:frame}, that we are able to get an optimum accuracy for a specific frame number which in turn makes the task easy of deciding the number of classes since it depends on $k$. 

Finally, we showcase some qualitative results for the Cambridge Landmark scenes as well as for the 7 scenes dataset. It can demonstrated from those results that not only is the learned anchor point different from the nearest one but also better, in terms of camera angle and in avoiding occlusion.

%
%
%
%
%
\vspace{-2.2em}
\section{Conclusion}
We propose a novel approach for solving the visual relocalization problem, inspired by how humans estimate their location, by observing suitable landmarks. We model it as a multi task problem of classification and relative offset regression. We propose a deep learning architecture and loss function which automatically discovers the anchor point relative to which accurate offset estimates can be predicted. We do not require each image to be tagged with the relevant landmark to train the classification part. Through our experiments, we show that our method  achieves an improvement over PoseNet and its extensions in all scenes of the Cambridge Landmarks dataset as well as the indoor scenes of the 7 Scenes dataset. We achieve 1.5m and 4$^\circ$ in localization performance in 4 out of the 6 outdoor scenes in Cambridge Landmarks and 0.2m localization for the 7 indoor Scenes. Furthermore, our method outperforms simple replacement of the feature extractor followed by regression which further showcases the advantages of an anchor point classification and relative offset regression based method for the visual localization task.



\bibliography{bmvc_review.bbl}

\end{document}